\pdfoutput=1
\documentclass{article}
\usepackage{amsmath,epsfig}
\usepackage[preprint]{spconfa4}
\usepackage{cite}
\usepackage{amssymb}
\usepackage{mathrsfs}
\newcommand*{\argmax}{\operatornamewithlimits{argmax}\limits}

\usepackage{booktabs}

\usepackage{multirow}
\usepackage{makecell}
\usepackage{boldline}
\usepackage{pifont}
\usepackage[dvipsnames]{xcolor}

\let\OLDthebibliography\thebibliography
\renewcommand\thebibliography[1]{
  \OLDthebibliography{#1}
  \setlength{\parskip}{0pt}
  \setlength{\itemsep}{0pt plus 0.3ex}
}

\begin{document}\sloppy

% Example definitions.
% --------------------
\def\x{{\mathbf x}}
\def\L{{\cal L}}

% Title.
% ------
\title{A novel stereo matching pipeline with robustness and unfixed disparity search range}
%
% Address.
% ---------------
\name{Jiazhi Liu$^{\ast\dagger}$, Feng Liu$^{\ast\dagger}$ }
\address{$^{\ast}$State  Key  Laboratory  of  Information  Security, Institute of Information Engineering, Chinese Academy \\ of Sciences; $^{\dagger}$School  of  Cyber  Security, University  of  Chinese Academy of Sciences\\Email: \{liujiazhi, liufeng\}@iie.ac.cn
\thanks{This work was supported by the National Key R\&D Program of China with No. 2018YFC0806900,  Beijing Municipal Science \& Technology Commission with Project No. Z191100007119009, NSFC  No.61902397, NSFC No. U2003111 and NSFC No. 61871378.}}

\maketitle

\begin{abstract}
Stereo matching is an essential basis for various applications, but most stereo matching methods have poor generalization performance and require a fixed disparity search range. Moreover, current stereo matching methods focus on the scenes that only have positive disparities, but ignore the scenes that contain both positive and negative disparities, such as 3D movies. In this paper, we present a new stereo matching pipeline that first computes semi-dense disparity maps based on binocular disparity, and then completes the rest depending on monocular cues. The new stereo matching pipeline have the following advantages: It 1) has better generalization performance than most of the current stereo matching methods; 2) relaxes the limitation of a fixed disparity search range; 3) can handle the scenes that involve both positive and negative disparities, which has more potential applications, such as view synthesis in 3D multimedia and VR/AR. Experimental results demonstrate the effectiveness of our new stereo matching pipeline.
\end{abstract}
\begin{keywords}
Stereo matching, Cross-domain generalization, View synthesis
\end{keywords}
\section{Introduction}
Recently, ``metaverse" has attracted many people's attention for its unlimited possibilities in the future. Compared with current 2D display techniques, ``metaverse" additionally utilizes binocular disparity~\cite{didyk2011perceptual} that refers to the slight difference between the left and right retina images, making people enjoy more realistic and immersive experience.   Therefore, disparity estimation plays a basic role in the techniques of ``metaverse". Since we have plenty of available stereoscopic content (such as 3D movies), there is a clear need for methods that can estimate disparity from such stereo pairs. 

Given a position $\mathbf{p}$ in the 3D world space, if its projections into the rectified left and right views are $\mathbf{p}\textsuperscript{L}$ and $\mathbf{p}\textsuperscript{R}$, then their disparity is defined as $\boldsymbol{D}\textsuperscript{L}(\mathbf{p}\textsuperscript{L})=\boldsymbol{D}\textsuperscript{R}(\mathbf{p}\textsuperscript{R})=\mathbf{p}^\text{L}_x-\mathbf{p}^\text{R}_x$ where $\mathbf{p}^\text{L}_x$ and $\mathbf{p}^\text{R}_x$ denote the horizontal components of $\mathbf{p}\textsuperscript{L}$  and $\mathbf{p}\textsuperscript{R}$. 
Current state-of-the-art stereo matching methods are designed and trained to match only over positive disparity ranges because the prior methods 1) mainly focused on the applications of autonomous driving and robot navigation that only require positive disparities; 2) need plenty of data to supervise the network, but there is not enough labeled data for the scenes containing both positive and negative disparities; 3) usually have poor generalization performance, so these methods have difficulties in utilizing large amounts of labeled data from known datasets.  In this paper, we explore the above challenges and present a new stereo matching pipeline: first computing semi-dense disparity maps based on binocular disparity, and then estimating the rest based on monocular cues. 

In contrast, the traditional stereo matching pipeline~\cite{scharstein2002taxonomy} usually contains four steps: matching cost computation, cost aggregation, disparity computation, and refinement.  Recent learning-based stereo matching methods also follow the traditional pipeline and design various network architectures. Here we review matching cost computation and cost aggregation that prior methods mainly involve. 

Matching costs are calculated based on hand-crafted or learning-based feature maps extracted from the left and right images, which expresses how well a pixel in the reference image matches its candidates in the other. Feature descriptors in traditional stereo matching methods are hand-crafted, usually based on gradients~\cite{klaus2006segment,hosni2012fast} and mutual information~\cite{hirschmuller2007stereo}. The early deep-learning-based methods focus on designing feature extraction networks to characterize the left and right images~\cite{zbontar2016stereo,luo2016efficient}, and then utilize non-learned cost aggregation and refinement to determine the final disparity map. The recent end-to-end methods also design various Siamese-like  networks to extract features, such as GCNet~\cite{kendall2017end}, StereoNet~\cite{khamis2018stereonet}, PSMNet~\cite{chang2018pyramid}, and GANet~\cite{zhang2019ga}, and their feature extractors are integrated into the whole network and trained in an end-to-end manner. After obtaining matching costs, we intuitively want to take the candidate with the smallest matching cost (or the largest similarity) as the best-matched pixel, but researchers generally do not adopt the intuitive operation for the following reasons: 1) the left and right features have poor ability to characterize the texture-less or repetitive regions; 2) the single-visible pixels do not have corresponding pixels in the other image. Therefore, for these ticklish pixels, the prior methods usually estimate their disparities from their neighboring tractable pixels, which is the job of cost aggregation. 

Cost aggregation denotes we aggregate space context information in the cost volume, so that we can compute more reliable disparities. Most traditional and some learning-based methods aggregate matching costs in a 3D cost volume (\emph{disparity$\times$height$\times$width}). Cost aggregation in a 3D cost colume, including SGM~\cite{hirschmuller2007stereo}, CostFilter~\cite{hosni2012fast},  DispNet~\cite{mayer2016large}, and AANet~\cite{xu2020aanet}, resembles the votes of neighboring pixels. Recent learning-based methods, such as PSMNet~\cite{chang2018pyramid}, GANet~\cite{zhang2019ga}, and CFNet~\cite{shen2021cfnet} execute cost aggregation in a 4D cost volume (\emph{feature$\times$disparity$\times$height$\times$width}), because researchers believe that 4D cost volumes contain more information than the 3D form. 

Motivated by the neuroscience fact that binocular disparity is a low-level, pre-attentive cue of human depth cues, we propose a new stereo matching pipeline that highlights the role of feature extractors. Different from the traditional stereo matching pipeline~\cite{scharstein2002taxonomy}, our proposed pipeline no longer needs cost aggregation: It first computes accurate semi-dense disparity maps directly from feature maps, and then generates the final disparity maps by disparity completion networks. 

In summary, our proposed stereo matching pipeline has the following advantages: 
    1) It has better generalization performance than most state-of-the-art stereo matching methods;
    2) It relaxes the limitation of a fixed disparity search range; 
    3) It can handle the scenes that involves both positive and negative disparities, whereas prior learning-based methods can only deal with the scenes with positive disparities;
    4 It does not store and process cost volumes at the inference time, so it needs less GPU memory.

\begin{figure*}[t!]
	\centering
	\includegraphics[width=0.95\linewidth]{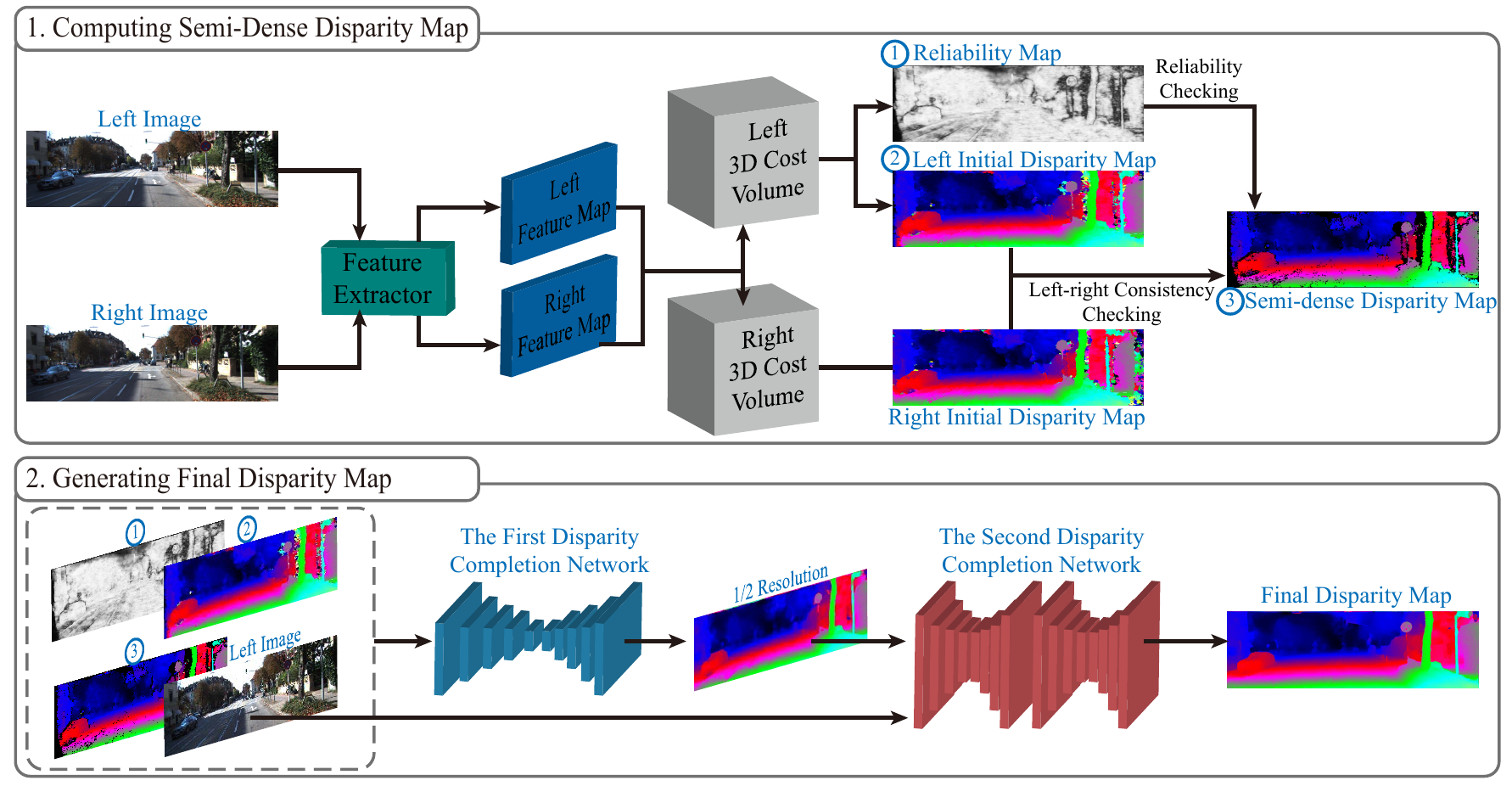}
	\caption{Overview of our proposed stereo matching pipeline at the inference time. The pipeline contains two stages: first computing semi-dense disparity maps directly from feature maps; then generating final disparity maps by two disparity completion networks. At the training time, we first train the first stage using the loss functions in Sec. \ref{sec:semi_loss}, and then the second stage. 
	}
	\label{fig:overview}
\end{figure*}

\section{Method}
Our goal is to compute the disparity map from a known rectified stereo pair. As mentioned above and shown in Figure \ref{fig:overview}, here we introduce the two steps of our stereo matching pipeline in detail. 

\begin{figure}[t!]
	\centering
	\includegraphics[width=0.8\linewidth]{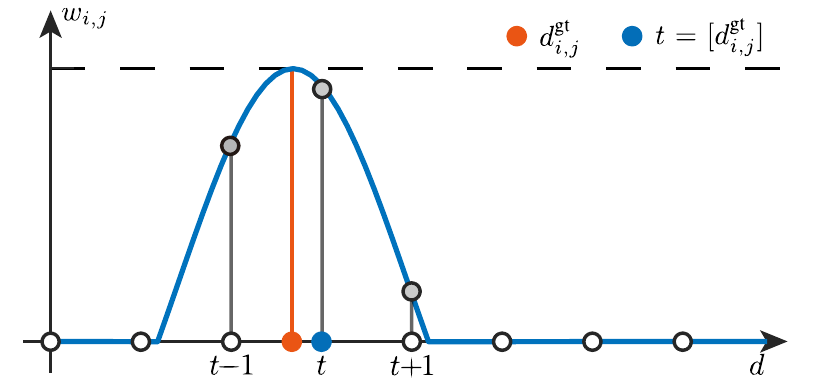}
	\caption{An example of weights of the double-visible-region loss.
	}
	\label{fig:w_function}
\end{figure}
\subsection{Computing Semi-dense Disparity Maps directly from Feature Maps}
Like prior methods, we adopt a shared feature extractor to extract the feature maps from the left and right images. Each feature map is a tensor with the size of $\alpha h\times\alpha w\times n$ where $h, w, n, \alpha$ represent image height, width, feature channels and the down-scaled scale. Here we introduce the symbol $\alpha$ to emphasize that current feature extractors generally extract feature maps at smaller resolution than the original image. Like AANet~\cite{xu2020aanet}, we utilize a full correlation to generate a  3D cost volume $\boldsymbol{C}=(c_{i,j}^{(d)})$ for pre-defined disparity candidates $\mathcal{D}=\{d\in\mathbb{Z}~|~\alpha d_\text{min}\leqslant d\leqslant \alpha d_\text{max}\}$. The 3D cost volume $\boldsymbol{C}$ is a tensor with the size of $\alpha h\times \alpha w\times m$ where $m=\alpha(d_\text{max}-d_\text{min})+1$, and each element $c_{i,j}^{(d)}$ records how well the position $(i/\alpha,j/\alpha)$ in the left image matches its disparity candidate $(i/\alpha,j/\alpha-d/\alpha)$ in the right image. 

After building a 3D cost volume, prior methods usually execute cost aggregation on the cost volume, so they need to store the volume at the whole inference process. In contrast, we generate an initial disparity map $D^\text{init}=(d_{i,j}^\text{init})\in[\alpha d_{\rm min},\alpha d_{\rm max}]^{\alpha h\times\alpha w}$ and the corresponding reliability map $R=(r_{i,j})\in[0,1]^{\alpha h\times\alpha w}$ directly from the cost volume, and thus there is no need to store the volume at the subsequent steps. The reliability map records how reliable a predicted disparity is, so we can obtain semi-dense disparity maps from initial disparity maps and reliability maps. 

We take the computation of disparity at the position $(i,j)$ as an example. We first convert the 3D cost volume $\boldsymbol{C}$ into the 3D probability volume $\boldsymbol{P}=(p_{i,j}^{(d)})$ by executing the softmax operation along the disparity dimension. 
Like the sub-pixel maximum a posteriori (MAP) ~\cite{tulyakov2018practical,nuanes2021soft}, we compute the initial  disparity map according to 
\begin{equation}
    d_{i,j}^\text{init}=\hat{d}_{i,j}+o_{i,j}
    \label{eq:disparity}
\end{equation}
where $\hat{d}_{i,j}=\mathop{\argmax}_{d\in\mathcal{D}}p_{i,j}^{(d)}$ denotes we chose the disparity candidate with the largest probability and $o_{i,j}$ is a refinement term to achieve sub-pixel precision. The refinement term $o_{i,j}$ is computed according to the possibilities at $\hat{d}_{i,j}-1$ and $\hat{d}_{i,j}+1$:
\begin{equation}
    o_{i,j}=
    \left.\sum_{\delta=-1}^{1}\delta\cdot p_{i,j}^{(\hat{d}_{i,j}+\delta)}\middle/\sum_{\delta=-1}^{1} p_{i,j}^{(\hat{d}_{i,j}+\delta)}\right. 
    \label{eq:refinement}
\end{equation}
where we regard $\sum_{\delta=-1}^{1} p_{i,j}^{(\hat{d}_{i,j}+\delta)}$ as the reliability $r_{i,j}$ of the initial disparity $d_{i,j}^\text{init}$. 

Finally, we filter the initial disparity map $D^\text{init}$ to  obtain a semi-dense disparity map by left-right consistency checking~\cite{hosni2012fast} and reliability checking (see Figure~\ref{fig:overview}). The latter checking refers to that the reliability of a predicted disparity has to be larger than the reliability threshold $\tau$. 

\subsubsection{Loss Function}
\label{sec:semi_loss}
Unlike prior stereo matching methods that train the feature extractor and rest modules in an end-to-end way, our method first trains the feature extractor since we have been highlighting its importance. The loss function contains three components: $\ell^\text{dvr}+\ell^\text{ur}+\ell^\text{trr}$.

\textbf{Double-visible-region loss} $\ell^\text{dvr}$.
For the position $(i,j)$ in the probability volume $\boldsymbol{P}$, we hope its maximal probability along the disparity dimension occurs at the ground truth disparity. Moreover, if the position $(i,j)$ is located in the occluded region, we cannot compute its disparity directly from feature maps since it has no corresponding pixel in the other. Therefore, we only execute this loss function in the double-visible region that can be derived from the ground-truth disparity map $D^\text{gt}=(d_{i,j}^\text{gt})$. Formally, the double-visible-region loss is defined as 
\begin{eqnarray}
    \ell\textsuperscript{dvr}=-\frac{1}{|\mathcal{V}|}\sum_{(i,j)\in\mathcal{V}}\left(\sum_{\delta=-1,0,1}w_{i,j}^{(t+\delta)}\cdot\log\left( p_{i,j}^{(t+\delta)}\right)\right)
    \label{eq:double_loss}
\end{eqnarray}
where $t=[d_{i,j}^\text{gt}]$ indicates the integral disparity closest to the ground truth and
the set $\mathcal{V}$ denotes all the pixels in the double-visible region. The coefficient $w^{(t+\delta)}_{i,j}$ is a weight function, and the closer to the ground truth a disparity is, the larger its weight will be (see Figure \ref{fig:w_function}). We give its detailed definition in the supplementary material. 

\textbf{Unreliable-region loss} $\ell^\text{ur}$. We hope that a predicted disparity has as low reliability as possible if it is located in the occluded region or its value diverges from the ground truth ($\geqslant 1$). Such the region is called the unreliable region $\mathcal{U}$.
The unreliable-region loss is defined as follows:
\begin{equation}
    \ell\textsuperscript{ur}=-\frac{1}{|\mathcal{U}|}\sum_{(i,j)\in\mathcal{U}}\log(1-r_{i,j}).
\end{equation} 

\textbf{To-refine-region loss} $\ell^\text{trr}$. If a predicted disparity $d_{i,j}^\text{init}$ is close to the ground truth, then we use this loss function to make the predicted disparity achieve more accurate sub-pixel precision. Formally, the loss function is 
\begin{equation}
    \ell^\text{trr}=\frac{1}{|\mathcal{T}|}\sum_{(i,j)\in\mathcal{T}}|d_{i,j}^\text{init}-d_{i,j}^\text{gt}|
\end{equation}
where $\mathcal{T}=\{(i,j)\in\mathcal{V}\big||d_{i,j}^\text{init}-d_{i,j}^\text{gt}|\leqslant 1\}$.

\subsection{Generating Final Disparity Maps by Disparity Completion Networks}
In the last subsection, we obtain a semi-dense disparity map, an initial disparity map, and a reliability map. %These maps generally have the $1/3$ resolutions relative to the original image. 
As shown in Figure \ref{fig:overview}, we utilize two disparity completion networks to complete hierarchically the semi-dense disparity and then acquire the final disparity map. The first disparity completion network is similar to the monocular depth estimation network~\cite{monodepth17}. It takes the initial disparity map, the semi-dense disparity map, the reliability map, and the reference color image as inputs, and outputs a full disparity map with the $1/2$ resolution. The second disparity completion network is similar to the refinement module in AANet~\cite{xu2020aanet}. It takes the full disparity map from the first network and the left image as inputs and outputs the final disparity map. % 画图

At the training time, the first disparity completion network also outputs the disparity maps with the resolution $1/4,1/8,1/16$. We train the two networks by supervising the five disparity maps with different resolutions. The corresponding loss function is the smooth L1 loss \cite{chang2018pyramid} that measures the difference between output disparity maps and the ground truths. More details are provided in the supplementary material.

\section{Experiments}
\subsection{Datasets}
We train the networks only in SceneFlow \cite{mayer2016large}, and evaluate them qualitatively  and quantitatively on the training sets of KITTI2012 \cite{geiger2012we}, KITTI2015 \cite{menze2015object}, Middlebury \cite{scharstein2014high}, which only contain positive disparities. We also use the dataset 3D Movie that includes negative disparities besides positive ones. 

\noindent\textbf{SceneFlow.}  SceneFlow is a large synthetic dataset containing 35,454 training and 4,370 test images with a resolution of $960\times540$, which have dense ground truth disparity maps.

\noindent\textbf{KITTI2012\&2015.} KITTI2012 and KITTI2015 are  real-world datasets with street views captured from a driving car, providing 394 stereo pairs of outdoor driving scenes with sparse ground-truth disparities for training, and 395 pairs for testing. 

\noindent\textbf{Middlebury.}
Middlebury is a small indoor dataset containing less than 50 stereo pairs with three different resolutions. Here we use its half version. 

\noindent\textbf{3D Movie.} We chose stereo images that contain both negative and positive disparities from the 3D movie "Big Buck Bunny" \copyright by Blender Foundation (https://peach.blender.org/). Their resolution is $1080\times1920$. 
The dataset does not have available ground truth disparities, so we just give a qualitative evaluation and visualized results. 

\subsection{Implementation Details}
We implemented the feature extractors in PyTorch  and using Adam ($\beta_1=0.9,\beta_2=0.999$) as optimizer. 
At training time we set the minimum disparity $d_{\text{min}}=0$ and the maximum disparity $d_{\text{max}}=320$, whereas at test time we adaptively adjusted the minimum and maximum disparity for different datasets. Specifically, we set $d_{\text{min}}=0,d_{\text{max}}=192$ for KITTI2012 and KITTI2015, $d_{\text{min}}=0,d_{\text{max}}=320$ for Middlebury, and set $d_{\text{min}}=-100,d_{\text{max}}=100$ for 3D Movie. 
We performed color normalization with ImageNet mean and standard deviation. During training, images were randomly cropped to the size $288\times576$, and their color was randomly augmented. We also used asymmetric data augmentation \cite{li2021revisiting}. 
We first trained the feature extractor for 18 epochs with a batch size $4$, and the learning rate started at 0.0005 and was decreased by half every 3 epochs after 6 epochs. After training the feature extractor, we then trained the rest module networks for 24 epochs, and the learning rate started at 0.001 and was decreased by half every 4 epochs after 8 epochs. All the models are trained and tested on one NVIDIA Tesla-P100 GPU. 

\begin{figure}[t!]
	\centering
	\includegraphics[width=\linewidth]{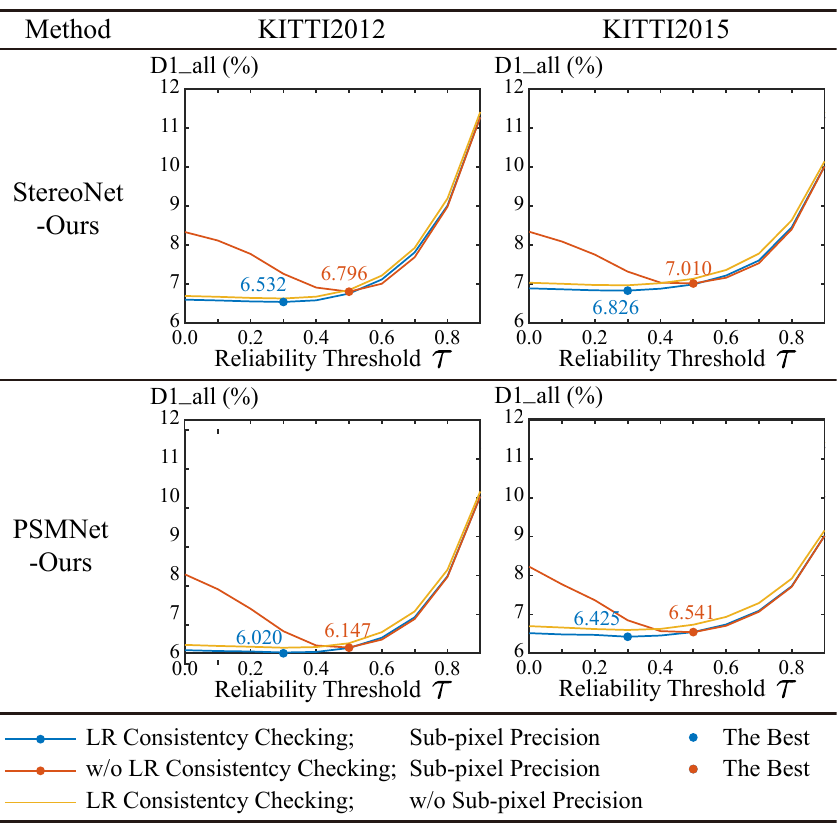}
	\caption{Ablation study results for left-right consistency checking,  sub-pixel precision, and  different reliability threshold values for reliability checking. ``StereoNet-Ours" denotes we utilized the feature extractor of StereoNet~\cite{khamis2018stereonet} in our proposed stereo matching pipeline. ``PSMNet-Ours" denotes we utilized the feature extractor of PSMNet~\cite{chang2018pyramid} in our proposed stereo matching pipeline. ``w/o sub-pixel precision" means we just use $\hat{d}_{i,j}$ in Eq.(2) as the initial disparity map. 
	}
	\label{fig:ablation}
\end{figure}
\subsection{Ablation Study}

We conducted experiments with several settings to evaluate our proposed stereo matching pipeline, including the usage of left-right consistency checking, sub-pixel precision,   different reliability threshold values for reliability checking, and different feature extractors (StereoNet~\cite{khamis2018stereonet} and PSMNet~\cite{chang2018pyramid}).  Left-right consistency checking and reliability checking work together to filter possibly-wrong disparities in the initial disparity map. Higher reliability threshold $\tau$ means more strict reliability checking. As shown in Figure \ref{fig:ablation}, if $\tau$ is lower than about 0.4, left-right consistency checking has a significant effect on the final result; otherwise, left-right consistency checking no longer work. 
Removing the sub-pixel precision module leads to a slight performance drop. 

For the methods ``StereoNet-Ours" and  ``PSMNet-Ours", we get the best result when using left-right consistency checking, reliability checking with $\tau=0.3$, and sub-pixel precision module. If we do not adopt left-right consistency checking, we can also obtain competitive results just depending on sub-pixel precision and reliability checking with $\tau=0.5$ (the blue dot v.s. the orange dot in Figure \ref{fig:ablation}).

\subsection{Evaluation}

\subsubsection{Cross-domain generalization performance}
\begin{table}[t!]
    \centering
    \footnotesize
    \begin{tabular}{c|ccc|c}\hlineB{2}
    \multirow{2}{*}{\makecell{Method}}& KITTI2012 & KITTI2015 & Middlebury & Time \\
    & D1\_all(\%)$\downarrow$ & D1\_all(\%)$\downarrow$ & bad2.0(\%)$\downarrow$ & (s)$\downarrow$ \\\hline
    \multicolumn{5}{c}{Traditional Methods}\\\hline
    CostFilter~\cite{hosni2012fast} & 21.7 & 18.9 & 40.5&$240^*$ \\
    SGM~\cite{hirschmuller2007stereo} & 7.1 & 7.6 & 25.2&$3.7^*$ \\
    \hline 
    \multicolumn{5}{c}{Domain-specific Methods}\\\hline
    PSMNet~\cite{chang2018pyramid}& 15.1&16.3 &25.1&0.337 \\
    GWCNet~\cite{guo2019group} & 20.2 & 22.7 & 34.2&$0.32^*$ \\
    AANet~\cite{xu2020aanet} & 18.3 & 12.2 & 31.0&0.123 \\\hline 
    \multicolumn{5}{c}{Domain-invariant Methods}\\\hline
    DSMNet~\cite{zhang2020domain} &6.2 &6.5&\textbf{13.8}&$1.6^*$ \\
    STTR~\cite{li2021revisiting} &8.4 &7.7&oom&1.416 \\\hline
    \multicolumn{5}{c}{Our new stereo matching pipeline}\\\hline
    StereoNet-Ours& 6.5& 6.8& 22.8&\textbf{0.095}\\
    PSMNet-Ours& \textbf{6.0}& \textbf{6.4}& 21.2&0.198\\
    \hlineB{2}
    \end{tabular}
    \caption{Generalization performance evaluation.  ``oom" denotes ``out of memory". We tested the running time on a NVIDIA P100 GPU for the KITTI resolution $384\times1344$. The symbol ``*" in Time denotes that the running time value is taken from the stereo evaluation system of KITTI. 
    }
    \label{tab:genelization}
\end{table}
Current state-of-the-art methods can achieve impressive performance in some special datasets, but usually have poor generalization performance. Here we evaluate the generalization performance of our stereo matching pipeline. Specifically, all the methods are trained in the synthesized dataset SceneFlow~\cite{mayer2016large}, and tested in three real-world datasets KITTI2012 \cite{geiger2012we}, KITTI2015 \cite{menze2015object}, and Middlebury \cite{scharstein2014high}. As shown in Table \ref{tab:genelization}, our method outperforms traditional methods ~\cite{hosni2012fast,hirschmuller2007stereo}, and the learning-based  domain-specific methods \cite{chang2018pyramid,guo2019group,xu2020aanet} on three datasets. STTR~\cite{li2021revisiting} adopts the prevailing \emph{transformer} architecture and is designed for cross-domain generalization. Our method  outperforms STTR on two datasets. Note that STTR failed in the half-resolution version of Middlebury, because  it requires too much GPU memory. DSMNet~\cite{zhang2020domain} is also designed for cross-domain generalization, and our method can surpass it in two datasets. Besides, our method is far faster than the two domain-invariant methods STTR~\cite{li2021revisiting} and DSMNet~\cite{zhang2020domain}.

\begin{table}[t!]
    \centering
    \footnotesize
    \begin{tabular}{ccccc}\toprule
    \multirow{2}{*}{\makecell{Method}}& Unfixed & Sub-pixel & Negative & \multirow{2}{*}{\makecell{Robust}}   \\
    & Range & Precision & Disparities &  \\\midrule
    CostFilter~\cite{hosni2012fast} & {\color{ForestGreen}\ding{51}} & {\color{Maroon}\ding{55}} & {\color{ForestGreen}\ding{51}} & {\color{Maroon}\ding{55}} \\
    SGM~\cite{hirschmuller2007stereo} & {\color{ForestGreen}\ding{51}} & {\color{Maroon}\ding{55}} & {\color{ForestGreen}\ding{51}} & {\color{ForestGreen}\ding{51}} \\
    \midrule 
    PSMNet~\cite{chang2018pyramid}& {\color{YellowOrange}\ding{51}\ding{55}} & {\color{ForestGreen}\ding{51}} & {\color{Maroon}\ding{55}} & {\color{Maroon}\ding{55}} \\
    AANet~\cite{xu2020aanet} & {\color{Maroon}\ding{55}} & {\color{ForestGreen}\ding{51}} & {\color{Maroon}\ding{55}} & {\color{Maroon}\ding{55}} \\\midrule
    DSMNet~\cite{zhang2020domain} & {\color{YellowOrange}\ding{51}\ding{55}} & {\color{ForestGreen}\ding{51}} & {\color{Maroon}\ding{55}} & {\color{ForestGreen}\ding{51}} \\
    STTR~\cite{li2021revisiting} & {\color{ForestGreen}\ding{51}} & {\color{ForestGreen}\ding{51}} & {\color{Maroon}\ding{55}} & {\color{ForestGreen}\ding{51}} \\\midrule
    Ours& {\color{ForestGreen}\ding{51}} & {\color{ForestGreen}\ding{51}} & {\color{ForestGreen}\ding{51}} & {\color{ForestGreen}\ding{51}}\\\bottomrule
    \end{tabular}
    \caption{Qualitative evaluation of our method and prior methods. The symbol {\color{YellowOrange}\ding{51}\ding{55}} denotes in the column ``Unfixed Range" denotes that the performance will be unstable if we change their disparity search range~\cite{tulyakov2018practical}.
    }
    \label{tab:qualitative}
\end{table}

\begin{figure}[t!]
	\centering
	\includegraphics[width=0.9\linewidth]{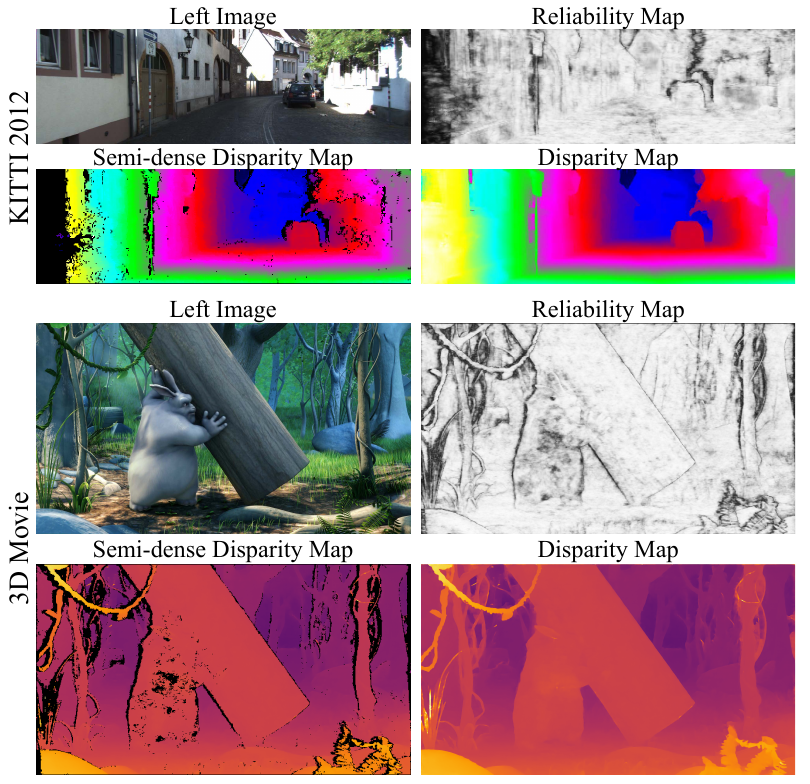}
	\caption{Visualized results. In reliability maps, bright colors means high reliability, whereas dark colors denotes low reliability.  
	}
	\label{fig:visualization}
	
\end{figure}
Table \ref{tab:qualitative} lists the qualitative evaluation results of our method and prior methods. For traditional methods, they cannot generate sub-pixel disparity maps because they use the scheme ``winner takes all". For learning-based methods, 1) if a method (such as AANet~\cite{xu2020aanet}) adopt cost aggregation on a 3D cost volume, its disparity search range is fixed and cannot be changed; 2) if a method (such as PSMNet~\cite{chang2018pyramid} and DSMNet~\cite{zhang2020domain}) executes cost aggregation on a 4D cost volume, its disparity range can be changed, but the performance is unstable if the range is changed. These learning-based methods are limited to supporting the scenes with only positive disparities, whereas our stereo matching pipeline can also support the scenes with both negative and positive disparities.

\subsubsection{Visualized Results}

We give an illustration of our method on KITTI2012 and 3D Movie.
 As Figure \ref{fig:visualization} shows, although our method is only trained in a synthesized dataset and over positive disparities, it still has a satisfying visualized result. Note that the stereo images of 3D Movie contain both positive and negative disparity, but it does not matter for our method. Figure \ref{fig:visualization} also shows reliability maps and semi-dense disparity maps. Surprisingly, the disparities of more than 80\% pixels can be computed accurately just depending on feature extractors, and most unknown disparities occur in the occluded region. 
 For more visualized results, please see the supplementary material.

\section{Conclusion}
In this paper, we have presented a new stereo matching pipeline that highlights binocular disparity of human depth cues, and no longer needs cost aggregation: It first computes accurate semi-dense disparity maps directly from feature maps, and then generates the final disparity maps by disparity completion networks. The new stereo matching pipeline 1) has superior generalization performance, 2) avoids the need to pre-specify a fixed disparity search range, and 3) can handle the scenes with both positive and negative disparities. We experimentally demonstrate that our stereo matching pipeline generalizes to different domains without fine-tuning and give visualized results on the scenes with both negative and positive disparities.

\bibliographystyle{IEEEbib}
\bibliography{main}

\end{document}